\documentclass{article}

\usepackage[
  a4paper,
  left=2.5cm,
  right=2.5cm,
  top=2cm,
  bottom=2.5cm,
  headsep=1cm,
  footskip=1cm
]{geometry}

\PassOptionsToPackage{hyphens}{url}
\usepackage{hyperref}

\usepackage{soulpos}
\usepackage{amsmath}
\usepackage{authblk}
\usepackage{graphicx}
\usepackage{fontawesome5}
\usepackage[T1]{fontenc}
\usepackage[utf8]{inputenc}
\usepackage[skins]{tcolorbox}
\usepackage[english]{babel}
\usepackage[autostyle=true]{csquotes}
\usepackage[nolist, nohyperlinks]{acronym}
\usepackage[capitalize, sort, english]{cleveref}
\usepackage[caption=false, font=footnotesize]{subfig}
 
\usepackage{tabularx}
\usepackage{multirow}
\usepackage{booktabs}
\newcolumntype{C}{>{\centering\arraybackslash}X}
\newcolumntype{R}{>{\raggedleft\arraybackslash}X}

\usepackage{tikz}
\usetikzlibrary{arrows.meta, positioning, calc}

\usepackage{xcolor}
\usepackage{listings}
\lstdefinelanguage{SPARQL}{
  morekeywords={SELECT,WHERE,VALUES,OPTIONAL,SERVICE,ORDER,BY,GROUP},
  sensitive=true,
  morecomment=[l]{\#},
  morestring=[b]",
}
\lstdefinelanguage{JSON}{
  morecomment=[l]{//},
  morestring=[b]",
  alsoletter={_},
  sensitive=true,
}
\lstdefinestyle{prompt}{
  language=JSON,
  escapeinside={(*@}{@*)},
  basicstyle=\ttfamily\small,
  commentstyle=\color[HTML]{000000},
  stringstyle=\color[HTML]{000000},
  keywordstyle=\color[HTML]{000000},
  showstringspaces=false,
  frame=single,
  breaklines=true,
  columns=fullflexible,
  upquote=true,
  string=[s]{"}{"},
  comment=[l]{:\ "},
  morecomment=[l]{:"},
  literate={→}{{$\rightarrow$\hspace{0.5em}}}1
}

\newcommand*\circled[1]{%
  \raisebox{0.2ex}{%
    \tikz[baseline=(char.base)]{
      \node[
        shape=circle,
        draw,
        inner sep=0.5pt,
        minimum size=10pt,
        font=\scriptsize
      ] (char) {#1};
    }%
  }%
}

\newcommand{\relation}[3]{%
  \enquote{#1} $\xrightarrow{\text{#2}}$ \enquote{#3}%
}

\usepackage[
    backend=biber,
    style=authoryear,
    natbib=true,
    maxbibnames=99,
    maxcitenames=2
]{biblatex}
\newcommand{\inception}{INCEpTION}

\usepackage{xcolor}
\definecolor{layerblue}{HTML}{4095cd}
\definecolor{layerorange}{HTML}{f8a069}
\definecolor{layerbrown}{HTML}{b17044}

\colorlet{layerbluelight}{layerblue!60}
\colorlet{layerorangelight}{layerorange!60}
\colorlet{layerbrownlight}{layerbrown!60}

\usepackage[skins]{tcolorbox}
\newtcbox{\custombox}[1][]{
    on line,
    nobeforeafter,
    tcbox raise base,
    boxrule=0.4pt,
    boxsep=2pt,
    top=0pt,
    bottom=0pt,
    left=0pt,
    right=0pt,
    arc=1pt,
    colframe=ul!,
    #1
}

\makeatletter
    \newcommand\prefixul[1][orange]{%
        \hspace{2pt}%
        \colorlet{ul}{#1}%
        \prefixul@
    }
    
    \ulposdef\prefixul@[xoffset-start=2pt, xoffset-end=-2pt]{%
        \ifulstarttype{0}%
            {\tcbset{ULsiderule/.append style={leftrule=0.4pt}}}%
            {\tcbset{ULsiderule/.append style={leftrule=0pt, sharp corners=west}}}%
        \ifulendtype{0}%
            {\tcbset{ULsiderule/.append style={rightrule=0.4pt, sharp corners=east}}}%
            {\tcbset{ULsiderule/.append style={rightrule=0pt, right=-2pt, sharp corners=east}}}%
        \tcbset{ULsiderule/.append style={colback=ul}}%
        \custombox[ULsiderule]{\vphantom{Ap}\rule{\ulwidth}{0pt}}%
    }

    \newcommand\suffixul[1][orange]{%
        \hspace{4pt}%
        \colorlet{ul}{#1}%
        \suffixul@%
    }
    
    \ulposdef\suffixul@[xoffset-start=2pt, xoffset-end=-2pt]{%
        \ifulstarttype{0}%
            {\tcbset{ULsiderule/.append style={leftrule=0pt, sharp corners=west}}}%
            {\tcbset{ULsiderule/.append style={leftrule=0pt, left=-2pt, sharp corners=west}}}%
        \ifulendtype{0}%
            {\tcbset{ULsiderule/.append style={rightrule=0.4pt}}}%
            {\tcbset{ULsiderule/.append style={rightrule=0pt, right=-2pt, sharp corners=east}}}%
        \tcbset{ULsiderule/.append style={colback=ul!10}}%
        \custombox[ULsiderule]{\vphantom{Ap}\rule{\ulwidth}{0pt}}%
    }
\makeatother

\newcommand{\content}[2]{%
    {\color{white}\prefixul[layerorange]{\textsf{#1}}}%
    {\color{black}\suffixul[layerorange]{#2}}\hspace{2pt}%
}

\newcommand{\entity}[2]{%
    {\color{white}\prefixul[layerblue]{\textsf{#1}}}%
    {\color{black}\suffixul[layerblue]{#2}}\hspace{2pt}%
}

\title{A Dataset for Named Entity Recognition and Relation Extraction from Art-historical Image Descriptions}

\author[1]{Stefanie Schneider}
\author[2]{Miriam Göldl}
\author[1]{Julian Stalter}
\author[3]{Ricarda Vollmer}

\affil[1]{Marburg University, Marburg, Germany}
\affil[2]{Ruhr University Bochum, Bochum, Germany}
\affil[3]{University of Munich, Munich, Germany}

\begin{document}

\begin{acronym}
\acro{AL}{\textit{Active Learning}}
\acro{AI}{\textit{Artificial Intelligence}}
\acro{CAS}{\textit{Common Analysis Structure}}
\acro{DFG}{\textit{German Research Foundation}}
\acro{NEL}{\textit{Named Entity Linking}}
\acro{FRAME}{\textit{Fine-grained Recognition of Art-historical Metadata and Entities}}
\acro{GLAM}{\textit{Galleries, Libraries, Archives, and Museums}}
\acro{IE}{\textit{Information Extraction}}
\acro{LLM}{\textit{Large Language Model}}
\acro{LOD}{\textit{Linked Open Data}}
\acro{NER}{\textit{Named Entity Recognition}}
\acro{NLP}{\textit{Natural Language Processing}}
\acro{RE}{\textit{Relation Extraction}}
\acro{UIMA}{\textit{Unstructured Information Management Architecture}}
\acro{XAI}{\textit{Explainable Artificial Intelligence}}
\end{acronym}

\maketitle

\begin{abstract}

This paper introduces \acsu{FRAME} (\acl{FRAME}), a manually annotated dataset of art-historical image descriptions for \ac{NER} and \ac{RE}.
Descriptions were collected from museum catalogs, auction listings, open-access platforms, and scholarly databases, then filtered to ensure that each text focuses on a single artwork and contains explicit statements about its material, composition, or iconography.
\acsu{FRAME} provides stand-off annotations in three layers:
a \textit{metadata layer} for object-level properties,
a \textit{content layer} for depicted subjects and motifs, and
a \textit{co-reference layer} linking repeated mentions.
Across layers, entity spans are labeled with $37$~types and connected by typed \ac{RE} links between mentions.
Entity types are aligned with Wikidata to support \ac{NEL} and downstream knowledge-graph construction.
The dataset is released as UIMA XMI \ac{CAS} files with accompanying images and bibliographic metadata, and can be used to benchmark and fine-tune \ac{NER} and \ac{RE} systems, including zero- and few-shot setups with \acp{LLM}.
\end{abstract}

\acresetall


\section{Background \& Summary}
\label{sec:background}


\ac{IE} converts unstructured text into a searchable, aggregable format, thereby enabling downstream tasks such as knowledge graph construction \citep{Repke21} and intra-collection linking \citep{DutiaS21}.
At its core, \ac{IE} decomposes two intertwined tasks: \ac{NER} and \ac{RE}.
\ac{NER} is, in essence, a problem of naming.
It asks where a text begins to refer to an entity (e.g., \textit{the Louvre}) and which category from a predefined inventory should be assigned to that entity, the \textit{entity type} (e.g., \textit{Museum}).
But in most settings---and certainly in art-historical discourse---recognizing isolated names is insufficient.
\ac{RE} therefore identifies and classifies semantic links between entity mentions: an artist in relation to a work of art (e.g., \textit{created by}), a work of art in relation to a religious figure (\textit{depicts}), or a place in relation to an institution (\textit{housed in}).
In practice, \ac{RE} presupposes a reliable \ac{NER}---as one cannot link what one has failed to identify---yet it also increases the difficulty of extraction: relations may be indirect, distributed across clauses or sentences, or often require shared domain knowledge.

However, models trained on contemporary, general-domain benchmarks \citep[e.g.,][]{SangM03, Weischedel13, DingXCWHXZL20} often underperform on historical writing.
There, vocabularies include domain-specific terminology, orthographic variation, unstable naming practices, and entity classes that are frequently not covered in standard inventories or knowledge bases \citep{VilainSL07}.
Curatorial texts further exacerbate these challenges, blending heterogeneous registers---object metadata, iconographic description, provenance, attribution, reception---while relying on long-distance dependencies and frequent co-references.
The result is a decrease in both quantitative and qualitative accuracy, with off-the-shelf systems being optimized to recognize categories that are not necessarily foregrounded in historical sources \citep{EhrmannHPRD24}.
Art history makes this mismatch unusually visible.
As shown by \citet{SierraK21}, applying general-purpose models on artwork-related corpora---digitized museum catalogs, for example---yields significant performance degradation.
Even state-of-the-art domain-adaptation and cross-domain methods---which typically leverage large quantities of unlabeled text---still fail to meet a basic prerequisite: they do not consistently recover the entity types that structure art-historical discourse in the first place \citep{ZhangXYLZ21}.
It is therefore unsurprising that \ac{NER} in this domain has often focused on narrower, better-scoped subproblems, such as detecting artwork titles \citep{JainMEK23}.
The practical implication is straightforward: progress in art-historical \ac{NER} cannot be secured by model choice alone; it depends on resources that explicitly formalize the domain's entity and relation inventory.
Existing datasets have advanced \ac{NER} and \ac{RE} in many domains, but their annotation schemes typically do not cover the entity and relation types that recur in art history, nor do they systematically encode the interplay between object-level metadata and depicted content.
As a result, deploying \ac{NLP} tools in digital art history usually demands substantial adaptation---yet a widely usable, expert-annotated reference dataset that captures the linguistic and conceptual structure of art-historical descriptions is still unavailable.
While \citet{GarciaV18} introduce, with SemArt, a multimodal dataset that provides images of paintings alongside short commentaries and structured attributes, it cannot function as a benchmark for token-level extraction, as it provides neither \ac{NER} spans and types nor explicit \ac{RE} annotations between mentions.
\citet{MuneraLDK24} address this gap from the perspective of \ac{NEL}.
Their MELArt dataset combines Artpedia descriptions with images, distinguishing  between \textit{visual} and \textit{contextual} sentences.
However, it also does not contain token-level \ac{NER} spans, a fine-grained type inventory, and explicit \ac{RE} annotations---components required for end-to-end \ac{IE} workflows.

\begin{figure*}[t!]
    \centering

    \begin{minipage}[t]{0.23\linewidth}
        \vspace{0pt}
        \begin{tcolorbox}[title=\circled{1} Image, top=0pt, bottom=0pt, left=0pt, right=0pt, coltitle=black, colframe=white]
            \includegraphics[width=\linewidth]{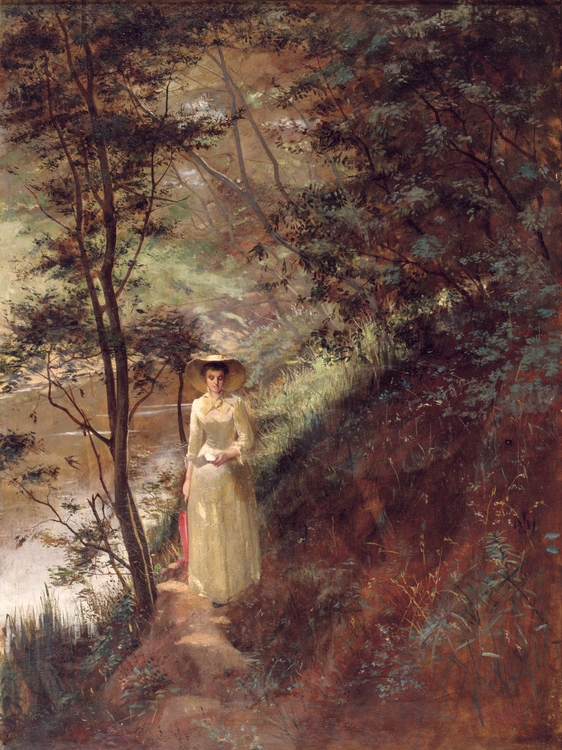}
        \end{tcolorbox}
    \end{minipage}
    \hfill
    \begin{minipage}[t]{0.74\linewidth}
        \vspace{0pt}
        \begin{tcolorbox}[title=\circled{2} Metadata, top=6pt, bottom=6pt, left=6pt, right=6pt, coltitle=black, colframe=white]
            {\small
            \textbf{Title:} The letter\hspace{12pt}
            \textbf{Creator:} Frederick McCubbin\hspace{12pt}
            \textbf{Date:} 1884\hspace{12pt}
            \textbf{Medium:} oil on canvas\hspace{12pt}
            \textbf{Source:} Wikipedia
            }
        \end{tcolorbox}
        \vspace{5pt}
        \begin{tcolorbox}[title=\circled{3} Text with Annotations, top=6pt, bottom=6pt, left=6pt, right=6pt, coltitle=black, colframe=white]
            {\small
            [$\ldots$] The \entity{Type of Work of Art}{painting} depicts a \content{Quality}{young} \content{Person}{woman} \content{Posture}{reading} a \content{Physical Object}{letter} \content{Posture}{walking} in the \content{Plant}{bush} alongside a \content{Geographical Feature}{stream}. [$\ldots$]
            }
        \end{tcolorbox}
    \end{minipage}

    \caption{
        Each record in the dataset includes (1)~the referenced artwork image, (2)~basic artwork metadata, and (3)~an art-historical text excerpt labeled with \ac{NER} and \ac{RE} annotations. 
    }
    \label{fig:annotation-example}
\end{figure*} 

To bridge this gap, we introduce \acsu{FRAME} (\acl{FRAME}), a manually annotated dataset of art-historical object descriptions with layered annotations spanning (i)~artwork metadata, (ii)~depicted figures, motifs, and visual elements, and (iii)~co-referenced entity mentions across the description.
Entity mentions are assigned to one of $37$~types that recur in art-historical writing---from artistic genres and religious figures to specialized terminology encountered on museum websites, auction houses, open-access platforms, and scholarly databases.
Entity types are aligned with Wikidata.
In addition, we annotate typed relations between entity mentions to support end-to-end extraction of structured knowledge.
Each description is paired with an image of the referenced artwork to support multimodal pipelines (\cref{fig:annotation-example}).
Our contributions are:
\begin{enumerate}
    \item An expert-annotated dataset of art-historical descriptions featuring a fine-grained entity taxonomy and typed relations, all of which are tailored to the domain.

    \item A multi-layer annotation that separates object metadata from depicted content, including co-reference links, which enable more accurate modeling of curatorial writing.

    \item An evaluation benchmark intended to support research in domain adaptation and knowledge graph construction for art history, making visible how well existing models generalize to the domain.
\end{enumerate}

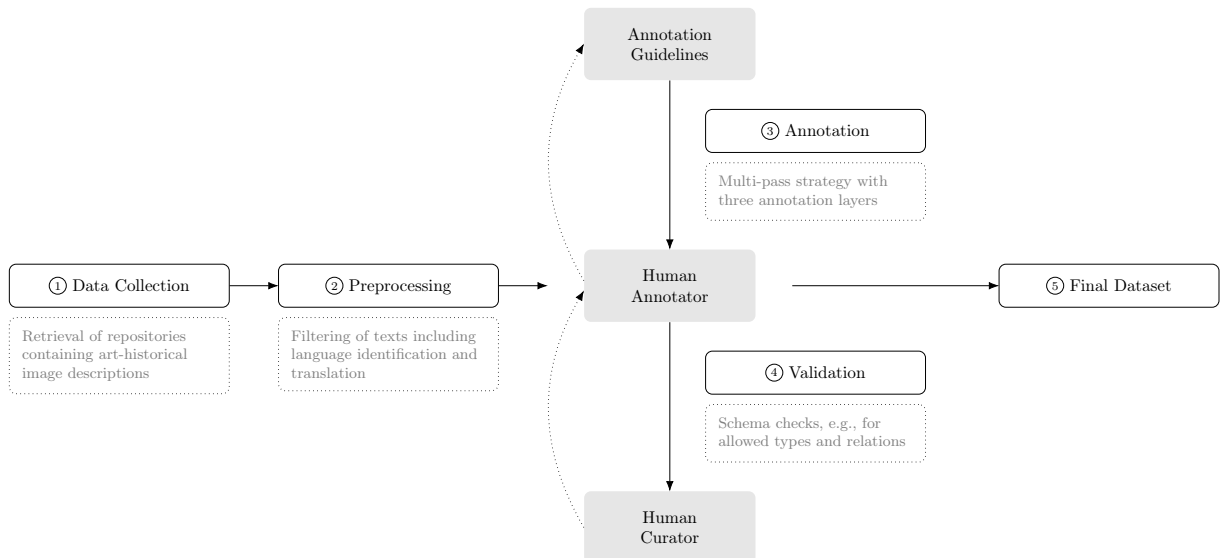
\begin{figure}[t!]
  \centering
  \resizebox{\textwidth}{!}{%
  \begin{tikzpicture}[
    node distance=1.0cm,
    >=latex,
    headline/.style={
      rectangle,
      rounded corners,
      draw=black,
      text width=4cm,
      minimum height=0.75cm,
      align=center,
      inner sep=7.5pt,
    },
    details/.style={
      rectangle,
      rounded corners,
      draw=black,
      dotted,
      text width=4cm,
      align=left,
      inner sep=7.5pt,
      font=\small,
      text=gray,
    },
    module/.style={
      rectangle,
      rounded corners,
      text width=3cm,
      minimum height=1.5cm,
      align=center,
      inner sep=7.5pt,
      fill=gray!20,
    },
    arrow/.style={-{Latex[length=2mm, width=1.5mm]}}
  ]

  \node[headline] (collection) {\circled{1} Data Collection};
  \node[details, below=0.2cm of collection] (collectionDetails) {Retrieval of repositories containing art-historical image descriptions};

  \node[headline, right=of collection] (preprocessing) {\circled{2} Preprocessing};
  \node[details, below=0.2cm of preprocessing] (preprocessingDetails) {Filtering of texts including language identification and translation};

  \node[module, right=1.75cm of preprocessing] (annotator) {Human\\Annotator};
  \node[module, above=3.5cm of annotator] (guidelines) {Annotation\\Guidelines};
  \node[module, below=3.5cm of annotator] (curator) {Human\\Curator};

  \node (annotationBlock) at ($(annotator)!0.5!(guidelines)+(3cm,0)$) [draw=none, inner sep=0] {
    \begin{tabular}{c}
      \begin{tikzpicture}[baseline=(v.base)]
        \node[headline] (v) {\circled{3} Annotation};
        \node[details, below=0.2cm of v] {Multi-pass strategy with three annotation layers};
      \end{tikzpicture}
    \end{tabular}
  };

  \node (validationBlock) at ($(annotator)!0.5!(curator)+(3cm,0)$) [draw=none, inner sep=0] {
    \begin{tabular}{c}
      \begin{tikzpicture}[baseline=(v.base)]
        \node[headline] (v) {\circled{4} Validation};
        \node[details, below=0.2cm of v] {Schema checks, e.g., for allowed types and relations};
      \end{tikzpicture}
    \end{tabular}
  };

  \node[headline, right=5cm of annotator] (dataset) {\circled{5} Final Dataset};

  \draw[arrow] (collection) -- (preprocessing);
  \draw[arrow, shorten >=0.75cm] (preprocessing) -- (annotator);
  \draw[arrow] (annotator) -- (curator);
  \draw[arrow] (guidelines) -- (annotator);
  \draw[arrow, shorten <=0.75cm] (annotator) -- (dataset);
  \draw[arrow, dotted, bend left=30] ([yshift=3pt]annotator.west) to (guidelines.west);
  \draw[arrow, dotted, bend left=30] (curator.west) to ([yshift=-3pt]annotator.west);
  \end{tikzpicture}
  }
  
  \caption{Our dataset creation process involves a modular, multi-stage pipeline, integrating both manual and automated components.}
  \label{fig:dataset-workflow}
\end{figure}

\section{Methods}
\label{sec:methods}


Our dataset is created using a modular, multi-stage pipeline, combining manual and automated components; \cref{fig:dataset-workflow} illustrates the overall workflow.
The following sections describe each stage in detail, specifying the tasks performed, the criteria for data selection and validation, and the rationale behind our methodological choices.
Our pipeline has two main objectives:
(i)~identifying texts that contain,  within a single cohesive paragraph, comprehensive descriptions of art-historical objects; and
(ii)~annotating these texts with entity types pertinent to art-historical research.

\subsection{Data Collection}
\label{sec:data-collection}

First, we surveyed and retrieved publicly accessible online repositories containing art-historical image descriptions.
In particular, we focused on:
(i)~museum websites with digital catalogs;
(ii)~auction houses with detailed lot listings;
(iii)~open-access platforms with collaboratively authored texts; and
(iv)~scholarly databases with peer-reviewed articles.
To extract content from both static and dynamically generated web pages, we developed a Python-based data acquisition routine.
For the asynchronous retrieval of static content, we employed the \textit{asyncio} library\footnote{\url{https://docs.python.org/3/library/asyncio.html} (accessed \today).}, while for dynamically generated or client-side-modified content, we utilized the \textit{Selenium} library.\footnote{\url{https://selenium-python.readthedocs.io/} (accessed \today).}
The workflow began with the identification of a suitable entry point, typically a base URL or a predictable URL pattern, such as a search or overview page that links to individual object records.
To traverse paginated results, we handled common URL parameters such as \enquote{offset}; in cases without explicit pagination, we heuristically inferred the maximum index to systematically capture object URLs.
Once the URLs had been compiled, we issued targeted requests to fetch individual object pages, parse the HTML structure, and extract the relevant textual content.
All harvested data were serialized and stored in JSONL format.
Using this methodology, we collected over $250{,}000$~textual entries from $13$~sources (\cref{tab:data-sources}).

The extracted texts varied widely in length and complexity: some consisted of brief summaries with only a few sentences, while others were full-length scholarly articles or book chapters.
To ensure the selection of texts with explicit references to specific artworks, we applied a two-stage filtering process.
First, we randomly sampled a subset of texts from the corpus.
Each text was manually reviewed and retained only if it satisfied both of the following criteria:
(i)~it explicitly describes a single art-historical object so that every annotated entity can be unambiguously linked to that object; and
(ii)~it contains at least one sentence that directly address the object or its creation, including references to materials, techniques, its composition, or iconography.
Even among qualifying texts, introductory and concluding sentences often digressed---summarizing the artist's broader practice, situating the work in art-historical context, or reiterating biographical details.
As our objective was to isolate content specifically about the artwork itself, we removed such extraneous sentences from the beginning or end of any text exceeding ten sentences; shorter texts were left unaltered.
The curated texts were then imported into \inception, an open-source annotation platform \citep{inception}.
Annotation was performed independently by four annotators, with each document assigned to two annotators.

\begin{table}[t!]
\caption{
    We collected over $250{,}000$~art-historical image descriptions from $13$~German and English sources.
    Each description contains at at least one sentence that directly addresses the object or its creation, including references to materials, techniques, its composition, or iconography.
    \vspace{10pt}
}
\label{tab:data-sources}

\footnotesize
\begin{tabularx}{\textwidth}{@{}Xll@{}}
\toprule
Name & URL & Language \\
\midrule
Austrian Gallery Belvedere          & \url{https://www.belvedere.at/en}            & DE \\
Bavarian State Painting Collections & \url{https://www.pinakothek.de/en}           & EN/DE \\
Berlin State Museums                & \url{https://www.smb.museum/en/}             & EN \\
Bonhams                             & \url{https://www.bonhams.com/}               & EN \\
Hamburger Kunsthalle                & \url{https://hamburger-kunsthalle.de/en}     & DE \\
Hampel Fine Art Auctions            & \url{https://www.hampel-auctions.com/}       & DE \\
Karl \& Faber Fine Art Auctions     & \url{https://www.karlundfaber.de/en/}        & DE \\
The Leiden Collection               & \url{https://www.theleidencollection.com/}   & EN \\
Lempertz                            & \url{https://www.lempertz.com/en/}           & EN/DE \\
Rijksmuseum                         & \url{https://www.rijksmuseum.nl/en}          & EN \\
Tate Gallery                        & \url{https://www.tate.org.uk/}               & EN \\
Van Ham                             & \url{https://www.van-ham.com/en/}            & DE \\
Wikipedia                           & \url{https://en.wikipedia.org/}              & EN/DE \\
\bottomrule
\end{tabularx}
\end{table}

\subsection{Data Annotation}
\label{sec:data-annotation}

Annotating art-historical texts for \ac{NER} is hard for reasons that are, at once, technical and epistemic.
The technical difficulty is well-known: the knowledge bases that typically underpin entity inventories tend to have uneven coverage of historical domains \citep{EhrmannHPRD24}.
This results in three recurring failure modes:
(i)~references to minor or lesser-known entities that are absent or underrepresented;
(ii)~entities whose names, borders, or institutional status have changed over time, or that exist only within a specific historical period; and
(iii)~ambiguous names that are shared by multiple entities across cultural and temporal contexts.
In art history, these issues are exacerbated by the continual renegotiation of titles, attributions, and the scholarly significance of objects.

The more fundamental difficulty, however, is not what \textit{can} be annotated, but what \textit{should} be annotated.
In search- and retrieval-oriented settings, the decisive question is which entity types actually function as meaningful access points to images and their descriptions.
Our tagset therefore focuses on
(i)~the intrinsic characteristics of the artwork as a historical artifact and
(ii)~the entities and concepts that structure its depiction.
We operationalize this distinction through a two-layer scheme:
in the \textit{metadata layer}, we prioritize entity types that describe the artwork as an object (its material and production context);
in the \textit{content layer}, we annotate what the artwork represents (its figures and iconographic motifs).
A third, \textit{co-reference layer}, links repeated mentions of the same entity---for instance, when a person is introduced by name and subsequently referred to by a pronoun.
The following sections detail the entity types selected for our tagset (\cref{sec:entity-type-selection}), the annotation procedure (\cref{sec:annotatation-procedure}), and the challenges---conceptual as well as practical---that emerged during the process (\cref{sec:annotation-challenges}).

\subsubsection{Entity Type Selection}
\label{sec:entity-type-selection}

To identify entities that are potentially useful for automatic extraction in search and retrieval contexts, we queried the Wikidata SPARQL endpoint and first extracted art-historical objects classified as either \enquote{visual artwork} (Wikidata item \texttt{Q4502142}) or \enquote{artwork series} (\texttt{Q15709879}).\footnote{\url{https://query.wikidata.org} (last accessed on \today). See \Cref{appendix:sparql-queries} for the respective SPARQL queries.}
We then analyzed the properties most frequently used to describe these objects and treated high-frequency properties as pragmatic indicators of what counts as salient, describable structure in the domain.
We selected Wikidata for three reasons.
First, it provides broader and more internally consistent coverage of objects than many specialized institutional databases (e.g., those maintained by the Metropolitan Museum of Art),\footnote{\url{https://www.metmuseum.org} (last accessed on \today).} which are necessarily shaped by institutional scope, regional focus, and curatorial priorities.
Second, unlike distributed aggregation platforms such as Europeana,\footnote{\url{https://www.europeana.eu} (last accessed on \today).} where reproductions and near-duplicates must be filtered out at scale, Wikidata is actively maintained and continuously updated, reducing the number of duplicates and thereby improving reliability.
Third, and most importantly, Wikidata is natively embedded within the broader \ac{LOD} ecosystem.
This matters: entity types derived from its schema remain directly linkable to a widely used knowledge graph, enabling interoperable downstream \ac{NER} and retrieval pipelines.

\begin{table}[t!]
\caption{
    The Wikidata properties most frequently associated with art-historical objects are shown with their relative frequency and an indicator of whether each property was included in the tagset.
    \vspace{10pt}
}
\label{tab:wikidata-properties}

\footnotesize
\begin{tabularx}{\textwidth}{@{}lXrr@{}}
\toprule
Property       & Property Label                      & Frequency & Include \\
\midrule
\texttt{P31}   & instance of                         &  7.82\,\% & yes \\
\texttt{P195}  & collection                          &  5.75\,\% &  no  \\
\texttt{P18}   & image                               &  5.57\,\% &  no  \\
\texttt{P276}  & location                            &  5.55\,\% &  no  \\
\texttt{P217}  & inventory number                    &  5.20\,\% &  no  \\
\texttt{P170}  & creator                             &  4.71\,\% & yes \\
\texttt{P571}  & inception                           &  4.39\,\% & yes \\
\texttt{P186}  & made from material                  &  3.44\,\% & yes \\
\texttt{P973}  & described at URL                    &  3.41\,\% &  no  \\
\texttt{P1476} & title                               &  3.34\,\% & yes \\
\texttt{P2048} & height                              &  3.19\,\% &  no  \\
\texttt{P2049} & width                               &  3.07\,\% &  no  \\
\texttt{P6216} & copyright status                    &  2.13\,\% &  no  \\
\texttt{P136}  & genre                               &  2.06\,\% & yes \\
\texttt{P17}   & country                             &  1.88\,\% &  no  \\
\texttt{P131}  & located in the administrative territorial entity & 1.65\,\% & no \\
\texttt{P180}  & depicts                             &  1.59\,\% & yes \\
\texttt{P625}  & coordinate location                 &  1.47\,\% &  no  \\
\texttt{P495}  & country of origin                   &  1.32\,\% &  no  \\
\texttt{P373}  & Commons category                    &  1.16\,\% &  no  \\
%
\bottomrule
\end{tabularx}
\end{table}

\cref{tab:wikidata-properties} lists the properties most frequently associated with art-historical objects in Wikidata, alongside their relative frequencies and a marker indicating whether we included them in our tagset.
The distribution shows that \enquote{instance of} (\texttt{P31}) is by far the most common property ($7.82$\,\%), as it is crucial for identifying the object type---e.g., distinguishing between a painting and an etching.
Other high-frequency properties are less useful for our purposes: \enquote{collection} (\texttt{P195}, $5.75$\,\%) and \enquote{location} (\texttt{P276}, $5.55$\,\%) primarily encode curatorial and institutional context; they rarely improve retrieval based on depicted content or intrinsic object characteristics, and were therefore excluded.
The same holds for technical identifiers such as \enquote{inventory number} (\texttt{P217}, $5.20$\,\%) and external links.
By contrast, we included properties that capture intrinsic, art-historically salient characteristics, such as \enquote{creator} (\texttt{P170}, $4.71$\,\%), \enquote{inception} (\texttt{P571}, $4.39$\,\%), \enquote{made from material} (\texttt{P186}, $3.44$\,\%), and \enquote{title} (\texttt{P1476}, $3.34$\,\%).
Together with \enquote{instance of,} these properties encode authorship, dating, and materiality---central dimensions for art-historical interpretation and comparison.
We also included properties that more directly target descriptive content.
Depiction-level information is most often given under \enquote{depicts} (\texttt{P180}, $1.59$\,\%), \enquote{main subject} (\texttt{P921}, $0.58$\,\%), and \enquote{depicts Iconclass notation} (\texttt{P1257}, $0.25$\,\%).
The most frequent top-level classes in this cluster include humans and mythological figures ($45.63$\,\%), with smaller percentages (each under $2$\,\%) for biological entities (plants, animals, taxa), built structures (such as churches or monuments), geographical features (rivers, mountains, settlements), and artistic or thematic concepts (iconographic motifs).
In Panofsky's terms, these classes correspond to the level of pre-iconographical description and iconographical analysis, where the focus shifts from identifying the object as a physical artifact to recognizing the figures, actions, and motifs depicted \citep{Panofsky06}.
In other words: these classes enable a transition from cataloging \enquote{what an object is} to indexing \enquote{what an object shows}---a prerequisite not only for humanistic interpretation but also for computational retrieval systems that aspire to operate on iconographic and, potentially, iconological levels of art-historical meaning.

\begin{table}[t!]
\caption{
    The tagset comprises $37$ entity types distributed across two annotation layers---the \textit{metadata layer} and the \textit{content layer}---with an additional \textit{co-reference layer} used to link mentions of the same entity.
    For each type, we provide a brief definition and, where applicable, the corresponding Wikidata item.
    Types that occur in both layers are annotated independently for each layer.
    \vspace{10pt}
}
\label{tab:entity-types}

\footnotesize
\begin{tabularx}{\textwidth}{@{}llXl@{}}
\toprule
Layer & Entity Name & Description & Wikidata Item \\
\midrule
\textit{Metadata} & Art Genre & \textit{form of art in terms of a medium, format, or theme} & \texttt{Q1792379} \\
 & Art Material & \textit{substance, raw ingredient, or tool that is utilized by an artist  [\ldots]} & \texttt{Q15303351} \\
 & Art Movement & \textit{tendency or style in art with a specific common philosophy  [\ldots]} & \texttt{Q968159} \\
 & Artistic Technique & \textit{method by which art is produced} & \texttt{Q11177771} \\
 & Type of Work of Art & \textit{type of art work based on shared characteristics, functions,  [\ldots]} & \texttt{Q116474095} \\
 & Work of Art & \textit{aesthetic item or artistic creation} & \texttt{Q838948} \\
 & Point in Time & \textit{position of a particular instant in time} & \texttt{Q186408} \\
 & Start Time & \textit{infimum of a temporal interval} & \texttt{Q24575110} \\
 & End Time & \textit{time that some temporal entity ceases to exist} & \texttt{Q24575125} \\
 & Person & \textit{being that has certain capacities or attributes constituting  [\ldots]} & \texttt{Q215627} \\
\midrule
\textit{Content} & Artistic Theme & \textit{theme or subject in a work of art} & \texttt{Q1406161} \\
 & Composition & \textit{placement or arrangement of visual elements in a work of art}
 & \texttt{Q462437} \\
 & Work of Art & \textit{aesthetic item or artistic creation} & \texttt{Q838948} \\
 & Concept & \textit{semantic unit understood in different ways, e.g., as mental  [\ldots]} & \texttt{Q151885} \\
 & Rhetorical Device & \textit{technique or strategy that a person uses with the goal of [\ldots]} & \texttt{Q1762471} \\
 & Emotion & \textit{biological states associated with the nervous system} & \texttt{Q9415} \\
 & Quality & \textit{distinguishing feature} & \texttt{Q185957} \\
 & Color & \textit{characteristic of visual perception} & \texttt{Q1075} \\
 & Point in Time & \textit{position of a particular instant in time} & \texttt{Q186408} \\
 & Season & \textit{section of a year} & \texttt{Q10688145} \\
 & Person & \textit{being that has certain capacities or attributes constituting  [\ldots]} & \texttt{Q215627} \\
 & Mythical Character & \textit{character from mythology} & \texttt{Q4271324} \\
 & Religious Character & \textit{character of a religious work, alleged to be historical} & \texttt{Q18563354} \\
 & Anatomical Structure & \textit{entity with a single connected shape} & \texttt{Q4936952} \\
 & Occupation & \textit{label applied to a person based on an activity they participate in} & \texttt{Q12737077} \\
 & Posture & \textit{physical configuration that a human can take} & \texttt{Q8514257} \\
 & Architectural Structure & \textit{human-designed and -made structure} & \texttt{Q811979} \\
 & Geographical Feature & \textit{components of planets that can be geographically located} & \texttt{Q618123} \\
 & Mythical Location & \textit{place that only exists in myths, legends, and folklore} & \texttt{Q3238337} \\
 & Religious Location & See Mythical Location. & \\
 & Physical Location & \textit{position of something in space} & \texttt{Q17334923} \\
 & Physical Surface & \textit{two-dimensional boundary of three-dimensional object} & \texttt{Q3783831} \\
 & Animal & \textit{kingdom of multicellular eukaryotic organisms} & \texttt{Q729} \\
 & Mythical Animal & \textit{creature in mythology and religion} & \texttt{Q24334299} \\
 & Food & \textit{any substance consumed to provide nutritional support for  [\ldots]} & \texttt{Q2095} \\
 & Physical Object & \textit{singular aggregation of substance(s), with overall properties  [\ldots]} & \texttt{Q223557} \\
 & Plant & \textit{living thing in the kingdom of photosynthetic eukaryotes} & \texttt{Q756} \\
\bottomrule
\end{tabularx}
\end{table}

These observations guided our selection of $37$ entity types for annotation (\cref{tab:entity-types}).
For the \textit{metadata layer}, we included properties that describe the intrinsic characteristics of the artwork itself---such as \enquote{instance of} (\texttt{P31}), \enquote{creator} (\texttt{P170}), \enquote{inception} (\texttt{P571}), and \enquote{made from material} (\texttt{P186}).
For the \textit{content layer}, we selected entity types derived primarily from the \enquote{depicts} hierarchy, supplemented by \enquote{main subject} (\texttt{P921}) and \enquote{depicts Iconclass notation} (\texttt{P1257}).
These properties capture figures, motifs, themes, and iconography, and thereby support more fine-grained retrieval and interpretation.
Separating \textit{metadata} and \textit{content} keeps object-level description and subject-level meaning analytically distinct without treating them as separate worlds.
Types that legitimately occur in both layers are annotated separately for each layer, reflecting their function in context: a title can serve as \textit{metadata} (a reference to the artwork) or as \textit{content} (a cited title within an interpretive comparison).

\begin{figure*}[t!]
    \centering

    \subfloat[Original text\label{fig:annotation-procedure-step-1}]{%
      \includegraphics[width=0.49\textwidth, trim={0 0 0 1.3cm}, clip]{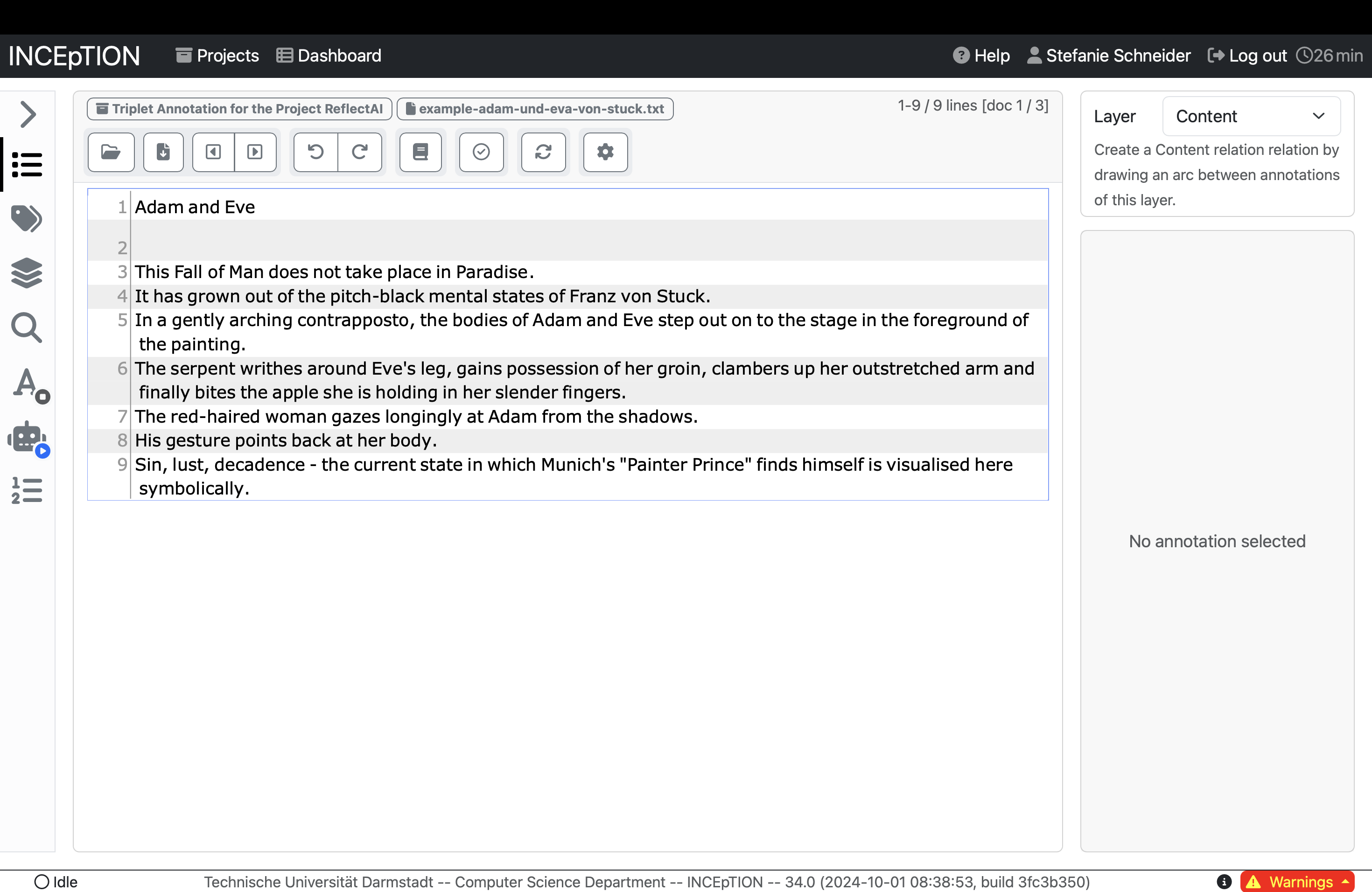}
    }
    \hfill
    \subfloat[Metadata layer\label{fig:annotation-procedure-step-2}]{%
      \includegraphics[width=0.49\textwidth, trim={0 0 0 1.3cm}, clip]{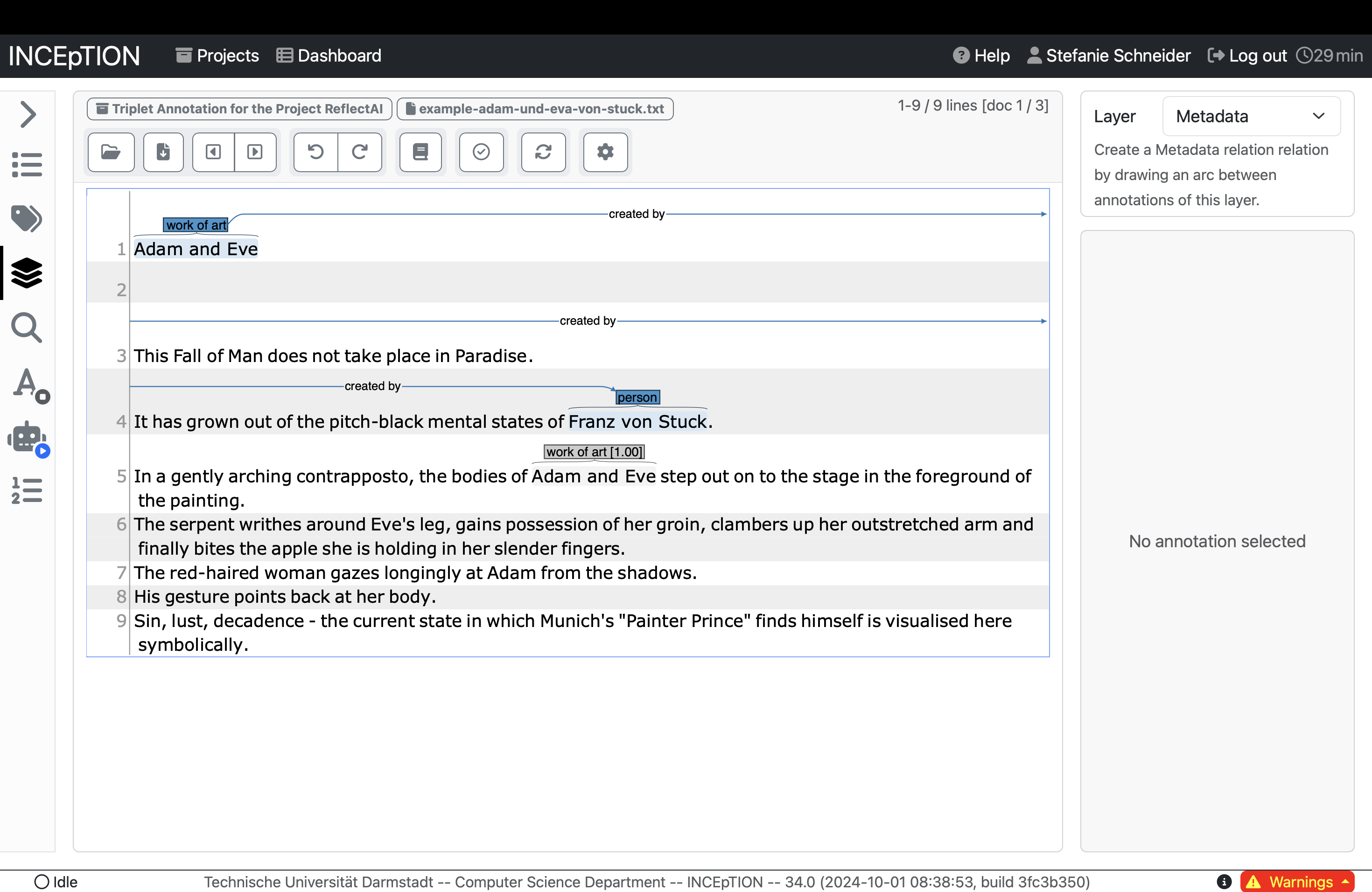}
    }

    \subfloat[Content layer\label{fig:annotation-procedure-step-3}]{%
      \includegraphics[width=0.49\textwidth, trim={0 0 0 1.3cm}, clip]{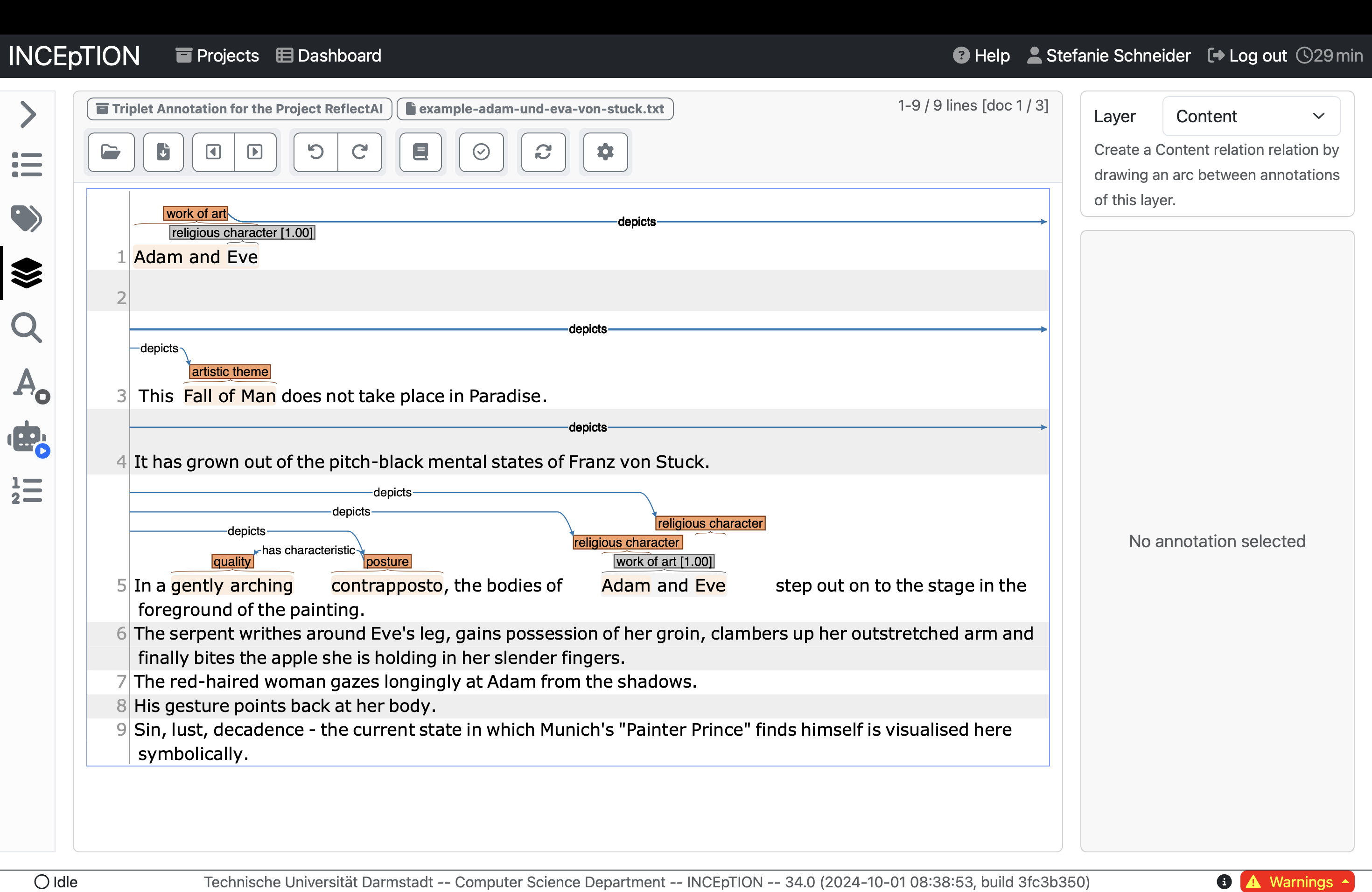}
    }
    \hfill
    \subfloat[Co-reference layer\label{fig:annotation-procedure-step-4}]{%
      \includegraphics[width=0.49\textwidth, trim={0 0 0 1.3cm}, clip]{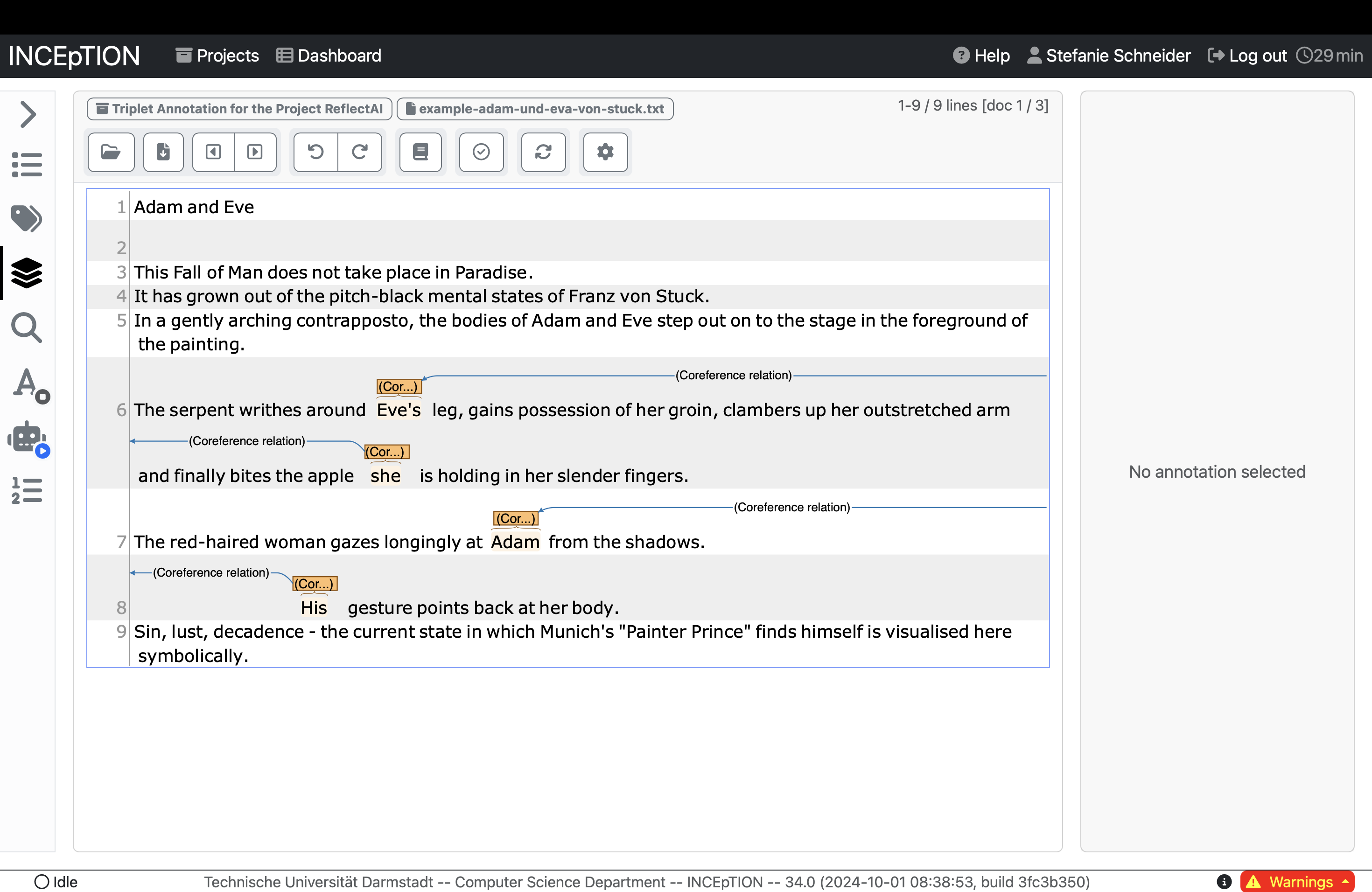}
    }

    \caption{
        Four-step annotation procedure.
        (a)~First, the text is read in full to obtain an overview, without creating annotations.
        (b)~Second, expressions belonging to the \textit{metadata layer} are annotated.
        (c)~Third, expressions in the \textit{content layer} are annotated.
        (d)~Fourth and finally, \textit{co-references} are annotated.
    }
\end{figure*}

\subsubsection{Annotation Procedure}
\label{sec:annotatation-procedure}

To ensure consistency in the annotation process---and make the interpretive work analytically legible---we implemented a multi-pass workflow in \inception.
By design, the procedure separates the afore-introduced three layers of the text:
(i)~a \textit{metadata layer},
(ii)~a \textit{content layer}, and
(iii)~a \textit{co-reference layer}.
This separation allows us to capture both descriptive and interpretive aspects of the texts without collapsing them into a single, difficult-to-audit annotation step.
Accordingly, each text was read at least four times, with each pass dedicated to one task in \inception:
\begin{enumerate}
    \item The text is first read in full to obtain an overview---without creating annotations---focusing on comprehension and on identifying potentially ambiguous passages (\cref{fig:annotation-procedure-step-1}).
    \item Then, all expressions belonging to the \textit{metadata layer} are annotated.
    Concretely, we (i)~identified candidate spans, (ii)~highlighted them, and (iii)~assigned the appropriate type from the drop-down menu (\cref{fig:annotation-procedure-step-2}).
    Once the relevant entities were in place, we created relations by linking entities with the mouse and selecting the appropriate relation label for each link.
    \item In a third reading, the same procedure is repeated for the \textit{content layer} (\cref{fig:annotation-procedure-step-3}).
    \item Finally, \textit{co-references} are annotated.
    As before, we marked candidate mentions as spans and linked co-referential expressions using the mouse-based relation interface (\cref{fig:annotation-procedure-step-4}).
\end{enumerate}
The annotation team consisted of four trained annotators.
Training followed a two-step process.
First, we distributed annotation guidelines, which includes an extensive markup file with definitions, decision rules, and illustrative examples.\footnote{See \url{https://doi.org/10.5281/zenodo.18724632}.}
Second, we held a joint training session via Zoom, during which three sample texts were annotated collaboratively.
This allowed the annotators to discuss any systematic ambiguities in the scheme, and to align their practical decisions with the shared interpretive commitments encoded in the guidelines.
Any corrections or clarifications arising from this exercise were incorporated into the final guideline document, ensuring that the annotation phase began from a common methodological baseline.

\subsubsection{Annotation Challenges}
\label{sec:annotation-challenges}

Art-historical prose shifts from what is \textit{seen} to what is \textit{known}, and from what is \textit{known} to what is \textit{meant}, often within the same sentence.
It is precisely this fluidity---between description, contextualization, and interpretation---that makes systematic annotation so challenging.
In practice, we encountered three recurring issues.

\begin{enumerate}
    \item The first issue is \textit{definitional}: what is \enquote{annotatable,} i.e., how to distinguish between (i)~what is explicitly depicted or materially present in the work of art, and (ii)~contextual information that remains visually absent from it?
    Art-historical writing often treats these registers as continuous, mirroring the gradual transition from pre-iconographical description to iconographical and iconological interpretation as described by Panofsky---a transition that is productive for art history, but problematic in terms of annotation \citep{Panofsky06}.
    Authors may incorporate personal associations, comparisons to related works, or broader disciplinary framing---language that sounds descriptive but fails the guideline requirement of being grounded in observable features.
    In Panofskian terms, such passages move beyond what is seen toward what is meant, even when they, linguistically, mimic descriptive statements \citep{Panofsky06}.
    This can be observed, for instance, in the Bavarian State Painting Collections' text on Johann Heinrich Füssli's \textit{Satan and Death with Sin Intervening} (1792/1802).\footnote{\url{https://www.sammlung.pinakothek.de/de/artwork/k2xnBqgxPd} (last accessed on \today).}
    The same problem recurs with expressions of atmosphere or innuendo, which frequently encode authorial judgment rather than directly observable content.

    \item A second issue is \textit{representational}: translating the nuance of art-historical description into the comparatively rigid logic of a fixed tagset.
    This is evident in multiple examples.
    For instance, art-historical discourse frequently refers to figures as carriers of social, professional, or religious identities---occupations (\textit{soldier}, \textit{king}), social roles (\textit{beggar}, \textit{mother}), or devotional or narrative functions (\textit{mourner}, \textit{saint}).
    Such attributions are central to iconographical identification, yet they often remain ambiguous in terms of annotation, as the available entity types do not consistently separate profession, social status, and narrative role.
    The tagset also exposes a structural limitation, as relational coherence and action are hard to encode.
    When texts describe figures interacting, gesturing, or participating in events, the annotation scheme largely reduces them to isolated entities and sparse relations.
    A related difficulty concerned formal description.
    Spatial organization, compositional placement, and pictorial syntax are often expressed through relational language.
    However, in the annotation, it was not always clear whether a phrase indicated a compositional term or merely provided narrative orientation---nor how to preserve that distinction in subject–predicate–object form (e.g., when to use \enquote{depicts} versus \enquote{contains}).
    Related is the issue of translating formal descriptions---such as spatial organization or compositional placement---into a fixed tagset and a small inventory of relations.
    Even when annotators agreed that compositional language belongs in the annotation, it was not always clear whether a phrase expressed composition (arrangement or placement) or merely provided narrative orientation; nor was it obvious how to encode this distinction in subject-predicate-object form (e.g., when to use the relation type \enquote{depicts} versus \enquote{contains}).
    The problem intensifies where multiple entities are described at once and relation direction and scope matter, as in Alonso il Sordillo del Arco's \textit{Abraham and the Three Angels} (1601--1700).\footnote{\url{https://www.hampel-auctions.com//a//Alonso-il-Sordillo-del-Arco-1635-Madrid-1704-zug.html?a=88&s=261&id=96857&q=96857} (last accessed on \today).}
    Here, compositional structure is articulated through phrases such as \enquote{on the right} and \enquote{on the left,} but the available relation logic encourages a sequential encoding that can flatten the described spatial syntax.
    
    \item A third issue is \textit{classificatory}: boundaries between entity types repeatedly are inconsistent.
    In the \textit{content layer}, for instance, the distinction between \enquote{architectural structure} (\textit{human-designed and -made structures}) and \enquote{geographical feature} (\textit{components of planets that can be geographically located}) is often difficult to sustain in borderline cases.
    In the \textit{metadata layer}, the difficulty shifts to overlaps between discursive categories---such as \enquote{art genre} and \enquote{art movement}.
    Although the guidelines distinguish between stylistic descriptors (e.g., \enquote{pre-impressionistic}) and collective movements (\enquote{Cubism}), art-historical prose does not always respect this distinction.
    The challenge, then, lies not in correctly identifying art-historically salient mentions, but in forcing them into discrete types under the constraints of a fixed scheme.
    The tension is especially pronounced for abstract or modern works, where perceptual phenomena organize the discourse, yet do not map cleanly onto existing entity types.
\end{enumerate}

\subsection{Data Validation}
\label{sec:data-validation}

Inter-annotator agreement is often treated as the default proxy for annotation quality \citep[see, e.g.,][]{ArtsteinP08}.
In this setup, the same document is first labeled by multiple annotators, with the labels then being collapsed into a single ground truth---typically by majority vote.
This is a pragmatic way of reducing noise.
But it also has the predictable consequence of privileging what is easiest to agree on.
Entities and relations that are rare, implicit, or contingent on domain knowledge---the \enquote{hard} cases---are precisely the ones most likely to be out-voted \citep{DavaniDP22}.
Voting also requires redundancy: every document must be annotated several times, and the procedure quietly assumes that a stable consensus is both attainable and meaningful for the task at hand.
To address these limitations, we transitioned from a one-shot aggregation scheme to a validation process that explicitly considers disagreement and uncertainty.
For the \ac{FRAME} dataset, we therefore adopted a multistage validation workflow \citep{MurphyDKJKJACK25}.
Rather than simply aggregating labels, we conducted a series of checks designed to make uncertain annotations visible and correctable.
Concretely, each document was independently annotated by two trained annotators in \inception.
Their outputs were then compared, and disagreements---competing entity type assignments or relations---were adjudicated by a senior curator with expertise in art history.
The curator could revise, merge, or extend annotations in light of the guidelines and the surrounding context.
Some disagreements, moreover, are not \enquote{mistakes,} but reflections of intentional representational choices---such as stacked tags when a single token legitimately has multiple functions.
For instance, in 2024, the Neumeister auction house offered a painting attributed to Frans Floris under the title \textit{Allegory of Death and Resurrection}.
Here, \enquote{allegory} is not a term that should be \textit{disambiguated} by voting, but rather be \textit{expanded}: by being tagged as both \enquote{art genre} in the \textit{metadata layer} and as \enquote{rhetorical device} in the \textit{content layer}.

\section{Data Record}
\label{sec:data-record}


The \ac{FRAME} dataset is publicly available on Zenodo under a Creative Commons Attribution 4.0 International (CC-BY-4.0) license: \url{https://doi.org/10.5281/zenodo.18724632}.
It is distributed as an \inception\ export, with one directory per document.
Each document directory contains (i)~the annotation package, \texttt{inception-document\{id\}.zip}, and (ii)~the corresponding curation state, \texttt{CURATION\_USER.ser}.
The annotation package includes a UIMA type system (\texttt{TypeSystem.xml}) and the curated annotations as a UIMA XMI \ac{CAS} file (\texttt{CURATION\_USER.xmi}).
The UIMA XMI CAS format provides a standardized representation of stand-off annotations over unstructured text: the \ac{CAS} is the \ac{UIMA} component that couples the primary text with its annotation layers, enabling interoperability with downstream \ac{NLP} pipelines.
Document directory names follow the normalized schema \texttt{\{title\}\_\{creator\}.txt}; both components are converted to lowercase, truncated to a maximum of $30$~characters, and sanitized by replacing non-alphanumeric sequences with underscores, collapsing repeated underscores, and removing leading or trailing underscores.
The referenced artwork images are stored in a ZIP file.
In addition, the dataset includes a CSV metadata file with bibliographic information for each document.
Each row corresponds to one document and provides the following fields: document directory, title, creator, artwork creation date, source institution, and a URL pointing to the corresponding record in the originating collection.

\section{Data Overview}
\label{sec:data-overview}


The dataset comprises $200$~image descriptions with an average length of $1{,}135.61$ characters; each description contains, on average, $48.44$~entities and $44.09$~relations.
Entity annotations are organized into three layers: the \textit{metadata layer} with $4{,}174$ entities and $1{,}277$ relations, the \textit{content layer} with $5{,}722$ entities and $5{,}931$ relations, and the \textit{co-reference layer} with $2{,}491$ entities and $1{,}609$ relations.
In the \textit{metadata layer}, the most frequent entity types are \enquote{person} ($520$~instances), \enquote{type of work of art} ($289$~instances), and \enquote{art genre} ($108$~instances);
in the \textit{content layer}, \enquote{quality} ($1{,}440$~instances), \enquote{composition} ($651$~instances), and \enquote{physical object} ($564$~instances) occur most often.\footnote{Statistics are based on version 1.0.0 of the dataset.}

\section{Technical Validation}
\label{sec:technical-validation}


\subsection{Data Acquisition}
\label{sec:data-acquisition}

To preserve stable and reproducible snapshots of the underlying sources, each retrieved webpage was archived in two representations:
(i)~the raw HTML, and
(ii)~a structured JSON record containing the extracted metadata and textual fields.
Because the corpus aggregates multilingual sources, all texts were first processed by an automatic language-identification module.
Texts not originally in English were then translated into English to standardize the downstream annotation and analysis pipeline.
For translation, we employed OpenAI's GPT-4o model, which consistently produced fluent and semantically faithful translations.\footnote{We also experimented with the 600-million-parameter distilled version of Meta's NLLB-200 model \citep{CostaCC22}. However, we observed that translations frequently contained ambiguous phrasing or omitted art-historical terminology.}
To assess translation quality, we conducted a small validation study on a stratified random sample of $20$ documents spanning different levels of art-historical specificity.
Bilingual reviewers evaluated each translation along three dimensions: semantic adequacy, preservation of named entities (e.g., artist names), and stylistic naturalness.

\subsection{Data Post-processing}
\label{sec:data-post-processing}

All annotations were validated against the guidelines using automated consistency checks.
These checks enforced a set of structural and typing constraints:
(i)~only entity and relation types defined by the guidelines were permitted;
(ii)~spans were not allowed to overlap within the same layer unless the guidelines explicitly permitted nesting;
(iii)~relations were required to connect type-compatible endpoints (e.g., \relation{work of art}{fabricated by}{artistic technique}, but not \relation{work of art}{fabricated by}{mythical character});
(iv)~every relation had to specify a non-empty type;
(v)~relations could not contain orphan endpoints;
(vi)~spans could not include leading or trailing whitespace;
(vii)~duplicate entities and relations were disallowed;
(viii)~each co-reference chain was required to include at least two mentions.
Violations were automatically flagged and resolved during the adjudication process, ensuring that the final dataset adhered strictly to the annotation guidelines.

\section{Usage Notes}
\label{sec:usage-notes}


The dataset is predominantly based on Western museum collections and the art-historical objects those institutions have long recognized as \textit{art}.
Objects situated in the domain of the arts and crafts---for instance, coins, weapons, tools, wooden carts with decorative carvings, and other utilitarian forms whose aesthetic value is inextricably linked to their function---are largely filtered out in advance.
This results in a structural preference for media that align with established museum taxonomies, such as painting or sculpture; however, it also leads to the under-representation of vernacular, functional, and hybrid artifacts that do not resolve into these categories.
A second bias is linguistic.
The dataset relies heavily on English translations of both primary and secondary materials, thereby importing Anglophone disciplinary conventions into the descriptions.
In several instances, fine-grained differences are obscured (e.g., between \enquote{panel} and \enquote{board,} or between \enquote{scene} and \enquote{episode}), with downstream effects on entity disambiguation.
This is particularly consequential for iconographic analysis: when an \textit{episode}---a discrete narrative unit with an implied sequence or liturgical framing---is rendered as a \textit{scene}, the text obviously loses information.

\section{Data Availability}
\label{sec:data-availability}


The \ac{FRAME} dataset is publicly available on Zenodo under a Creative Commons Attribution 4.0 International (CC-BY-4.0) license: \url{https://doi.org/10.5281/zenodo.18724632}.
It is distributed as an \inception\ export, with one directory per document.
Each directory contains (i)~the annotation package and (ii)~the corresponding curation state; the package includes the UIMA type system and the curated annotations as a UIMA XMI \ac{CAS} file.
Document directory names follow the normalized schema \texttt{\{title\}\_\{creator\}.xmi}.
The referenced artwork images are stored in a ZIP file.
In addition, the dataset includes a CSV metadata file with bibliographic information for each document: document name, title, creator, artwork creation date, source institution, and a URL pointing to the corresponding record in the originating collection.

\section{Code Availability}
\label{sec:code-availability}


\inception, the platform used for semantic text annotation, is released under the Apache License 2.0 and is available on GitHub: \url{https://github.com/inception-project/inception}.
To support reuse of the dataset, we provide the validation code under the GNU General Public License 3.0 at: \url{https://github.com/stefanieschneider/frame-toolset}.

\printbibliography

\section*{Acknowledgements}


We thank Edith Funfrock, Elias Neuhaus, and Theresa Zischkin for their valuable support with image annotation.
We also thank Hubertus Kohle, Ralph Ewerth, Eric Müller-Budack, and Matthias Springstein for their insightful discussions and helpful comments on the subject matter.

\section*{Author Contributions}


S.S. conceived, designed, and conducted the experiments; analyzed the data; supervised image annotation and curation; and ensured data quality.
M.G. and R.V. contributed to the annotation and curation process.
S.S., M.G., J.S., and R.V. co-authored the manuscript and reviewed, commented on, and approved the final version.

\section*{Competing Interests}


The authors declare no competing interests.

\section*{Funding}


This work was funded in part by the \acf{DFG} under project no. 510048106.

\appendix

\section{SPARQL Queries}
\label{appendix:sparql-queries}

We employ three SPARQL queries to extract relevant objects from Wikidata.
The first query restricts the search space to subclasses of either \enquote{visual artwork} (\texttt{Q4502142}) or \enquote{artwork series} (\texttt{Q15709879}).
The query is as follows:
\begin{lstlisting}
SELECT DISTINCT ?item WHERE {
  VALUES ?superClass { wd:Q4502142 wd:Q15709879 }
  ?item wdt:P279/wdt:P279 ?superClass .
}
\end{lstlisting}
Second, we retrieve objects that are instances (or subclasses) of a given superclass.
To be included, each object must provide at least a two-dimensional image (\texttt{P18}) or an official website (\texttt{P973}):
\begin{lstlisting}
SELECT ?item WHERE {
  VALUES ?superClass { wd:subclass_id }
  ?item wdt:P31/wdt:P279* ?superClass .

  FILTER (
     EXISTS { ?item wdt:P18  ?image }
  || EXISTS { ?item wdt:P973 ?url   }
  )
}
\end{lstlisting}
Finally, given the resulting identifiers, we retrieve (for each object) its associated properties, their values, and English labels.
Because this step is resource-intensive, the items are processed in batches of $50$ (denoted by \texttt{itemArray}) to reduce the likelihood of query time-outs:
\begin{lstlisting}
SELECT DISTINCT ?id ?property ?propertyLabel WHERE {
  VALUES (?id) { itemArray }

  ?id ?p ?statement .
  ?property wikibase:claim ?p   .

  SERVICE wikibase:label { bd:serviceParam wikibase:language "en" }
}
ORDER BY ?id ?property
\end{lstlisting}

\end{document}